\documentclass[]{elsart}

\usepackage{natbib}

\usepackage{epsfig}

\usepackage{amssymb}

\begin{document}


\begin{frontmatter}


\title{The Self-Organization of Speech Sounds}

\author{Pierre-Yves Oudeyer}  
\ead{py@csl.sony.fr}
\ead[url]{www.csl.sony.fr/$\sim$py}
\address{Sony CSL Paris, \\ 6, rue Amyot, \\ 75005 Paris, France}








\begin{abstract}
\noindent 

The speech code is a vehicle of language: it defines
a set of forms used by a community to carry information.
Such a code is necessary to support the linguistic
interactions that allow humans to communicate. 
How then may a speech code be formed prior to the 
existence of linguistic interactions?
Moreover, the human speech code is discrete and compositional,
shared by all the individuals of a community but different
across communities, and phoneme inventories are characterized by
statistical regularities. How can a speech code with these properties form?
 
We try to approach these questions in the paper,
using the ``methodology of the artificial''. We 
build a society of artificial agents, and detail a mechanism that
shows the formation of a discrete speech code without pre-supposing
the existence of linguistic capacities or of coordinated interactions.
The mechanism is based on a low-level model of
sensory-motor interactions. We show that the integration of certain very 
simple and non language-specific neural devices 
leads to the formation of a speech code that
has properties similar to the human speech code.
This result relies on the self-organizing properties of a generic
coupling between perception and production
within agents, and on the interactions between agents.
The artificial system helps us to develop better intuitions on how speech
might have appeared, by showing how self-organization
might have helped natural selection to find speech.

\end{abstract}


\end{frontmatter}

\section{The origins of language: a growing field of research}

A very long time ago, human vocalizations were inarticulate grunts. Now, humans speak. 
The question of how they came to speak is one of the most difficult that science has to tackle. 
After its ban from scientific inquiry during most of the 20th century, because of the Soci\'et\'e 
Linguistique de Paris declared in its constitution that is was not a scientific question, it is now again the centre of research of a 
growing scientific community. The diversity of the problems which are implied requires a high 
pluri-disciplinarity: linguists, anthropologists, neuroscientists, primatologists, psychologists
but also physicists and computer scientists belong to this community. Indeed, a growing number 
of researchers on the origins of language consider that a number of 
properties of language can only be explained by the dynamics of the complex interactions between 
the entities which are involved (the interaction between neural systems, the vocal tract, the ear, 
but also the interactions between individuals in a real environment). This is the contribution of the 
theory of complexity \citep{Nicolis77}, developed in the 20th century, which tells us
that there are many natural systems in which macroscopic properties can not be deduced 
directly from the microscopic properties.  This is what is called self-organization. Self-organization
is a property of systems composed of many interacting sub-systems, where the patterns and dynamics at
the global level are qualitatively different from the patterns and dynamics of the sub-systems. 
This is for example the case of the fascinating structures of termite nests \citep{Bonabeau99}, 
whose shape is neither coded 
nor known by the individual termites, but appears in a self-organized manner when termites interact. 
This type of self-organized dynamics is very difficult to grasp intuitively. The computer 
happens to be the most suited tool for their exploration and their understanding \citep{Steels97}. 
It is now an essential tool in the domain of human sciences 
and in particular for the study of the origins of language \citep{Cangelosi02}. One of the objectives of this paper is to 
illustrate how it can help our understanding to progress.

We will not attack the problem of the origins of language in its full generality, but rather we will focus 
on the question of the origins of one of its essential components : speech sounds, the vehicle and 
physical carriers of language.

\section{The speech code}

Human vocalization systems are complex. Though physically continuous acoustico-motor trajectories, 
vocalizations are cognitively discrete and compositional: they are 
built with the re-combination of units which are systematically re-used. These units are present at 
several levels \citep{Browman86}: gestures, coordination of gestures or phonemes, morphemes. 
While the articulatory space which defines the space of physically possible gestures is continuous, 
each language discretizes this space in its own way, and only uses a small and finite set
of constriction configurations when vocalizations are produced as opposed to using configurations
which span all the continuous articulatory space: this is what we call is the discreteness of the speech code.
While there is a great diversity across the 
repertoires of these units in the world languages, there are also strong regularities (e.g. the frequency 
of the vowel system /a, e, i, o, u/ as shown in \citep{Schwartz97}).

Moreover, speech is a conventional code. Whereas there are strong statistical regularities across 
human languages, each linguistic community possess its own way of categorizing sounds. 
For example, the Japanese do not hear the difference 
between the [r] of ``read'' and the [l] of ``lead''. How can a code, shared by all the members of a 
community, appear without centralised control? It is true that since the work of de Boer \citep{deBoer01}
or Kaplan \citep{Kaplan01}, we know how a new sound or a new word can propagate and be accepted in a 
given population. But this is based on mechanisms of negotiation which pre-suppose the existence 
of conventions and of linguistic interactions. These models are dealing with the cultural evolution of languages, but do 
not say much about the origins of language. Indeed, when there were no conventions at all, how 
could the first speech conventions have appeared?

\section{How did the first speech codes appear?}

It is then natural to ask where this organization comes from, and how a shared speech code could 
have formed in a society of agents who did not already possess conventions. Two types of answers 
must be provided. The first type is a functional answer: it establishes the function of sound systems, 
and shows that human sound systems have an organization which makes them efficient for achieving 
this function. This has for example been proposed by Lindblom \citep{Lindblom92} who showed that statistical 
regularities of vowel systems could be predicted by searching for the vowel systems with quasi-optimal 
perceptual distinctiveness. This type of answer is necessary, but not sufficient: it does not 
explain how evolution (genetic or cultural) may have found these optimal structures, and how a community 
may choose a particular solution among the many good ones. In particular, it is possible that 
``naive''  Darwinian search with random variations is not efficient enough for finding complex structures 
like those of speech: the search space is too big \citep{Ball01}. This is why a second type of answer is 
necessary: we have to account for how natural selection may have found these structures. A possible 
way to do that is to show how self-organization can constrain the search space and help natural selection. 
This may be done by showing how a much simpler system can self-organize spontaneously and form the 
structure we want to explain.

\begin{figure}[t!]
\centerline{\psfig{figure=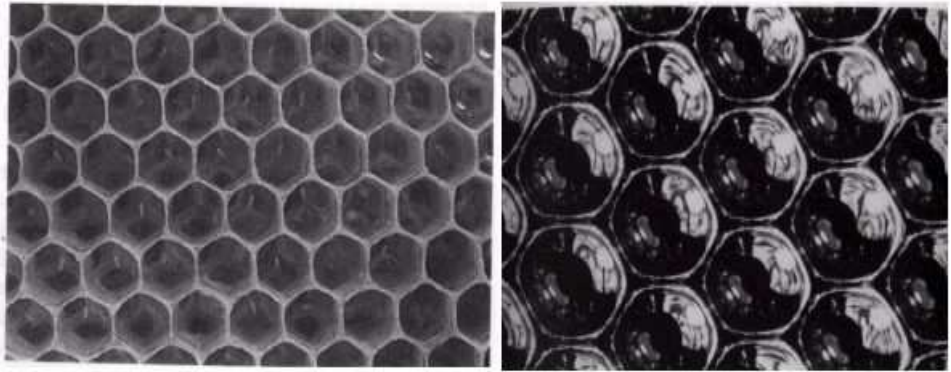,height=2in,width=5in,angle=0}}

\caption{\label{honeybee}
\small{The cells in the honey-bees nests (figure on the left) have a perfect hexagonal shape. Packed water bubbles
take spontaneously this shape under the laws of physics (figure on the right). This lead D'Arcy Thompson to think that
these same laws of physics might be of great help in the building of their hexagonal wax cells.
}} 
\vskip 0.5cm 
\end{figure} 

The structure of our argumentation about the origins of speech is the same as the one of D'Arcy Thompson 
\citep{Thompson32} about the explanation of hexagonal cells in honey-bees nests (see Figure \ref{honeybee}). 
The cells in the honey-bees nests have a perfect hexagonal shape. How did bees came to build 
such structures? A first element of answer appears if one remarks that the hexagon is the shape which 
necessitates the minimum amount of wax in order to cover a plane with cells of a given surface. So, the 
hexagon makes the bees spend less metabolic energy, and so they are more efficient for survival and 
reproduction than if they would build other shapes. One can then propose the classical neo-Darwinian 
explanation : the bees must have begun by constructing random shapes, then with random mutations 
and selections, more efficient shapes were progressively found, until one day the perfect hexagon was 
found.  Now, a genome which would lead a bee to build exactly hexagons must be rather complex and is 
really a needle in a haystack. And it seems that the classical version of the neo-Darwinian mechanism 
with random mutations is not efficient enough for natural selection to have found such a genome. So the 
explanation is not sufficient. D'Arcy Thompson completed it. He remarked that when wax cells, with a 
shape not too twisted, were heated as they actually are by the working bees, then they have approximately 
the same physical properties as water droplets packed one over the other. And it happens that
when droplets are packed, they spontaneously take the shape of hexagons. So, D'Arcy Thompson 
shows that natural selection did not have to find genomes which pre-program precisely the construction of
 hexagons, but only genomes who made bees pack cells whose shape should not be too twisted, and 
then physics would do the rest\footnote{This does not mean that nowadays honey bees have not a precise
innate hard wired neural structure which allows them to build precisely hexagonal shapes, as has been
suggested in further studies such as those of \citep{Frisch74}. The argument of D'Arcy Thompson just says that initially
the honey bees might have just relied on the self-organization of heated packed wax cells, which would
have lead them to ``find'' the hexagon, but later on in their evolutionary history, they might have
incorporated in their genome schemata for building directly those hexagons, in a process similar
to the Baldwin effect \citep{Baldwin96}, in which cultural evolution is replaced here by the self-organization
of coupled neural maps.}.
He showed how self-organized mechanisms (even if the term did not 
exist at the time) could constrain the space of shapes and facilitate the action of natural selection.
We will try to show in this paper how this could be the case for the origins of speech sounds.

Some work in this direction has already been developed in \citep{Browman00}, \citep{deBoer01}, 
and \citep{Oudeyer01b} concerning speech, and in \citep{Steels97}, \citep{Kirby01} or \citep{Kaplan01}, concerning lexicons
and syntax. These works provide an explanation of how a convention like the speech
code can be established and propagated in a society of contemporary human speakers.
They show how self-organization helps in the establishment of society-level conventions
only with local cultural interactions between agents.
But all these works deal rather with the cultural evolution of languages than with the origins of language.
Indeed, the mechanisms of convention propagation that they use necessitate already the
existence of very structured and conventionalised interactions between individuals. They
pre-suppose in fact a number of conventions whose complexity is already ``linguistic''.

Let us illustrate this point with the work of de Boer \citep{deBoer01}.
He proposed a mechanism for explaining how a society of agents may come
to agree on a vowel system. This mechanism is based on mutual imitations
between agents and is called the ``imitation game''.
He built a simulation in which agents were given a model
of the vocal tract as well as a model of the ear. Agents played a game
called the imitation game. Each of them had a repertoire of
prototypes, which were associations between a motor program and its
acoustic image. In a round of the game, one agent called the speaker,
chose an item of its repertoire, and uttered it to the other agent,
called the hearer. Then the hearer would search in its repertoire for the
closest prototype to the speaker's sound, and produce it (he
imitates). Then the speaker categorizes the utterance of the hearer
and checks if the closest prototype in its repertoire is the one he 
used to produce its initial sound. He then tells the hearer whether
it was ``good'' or ``bad''. All the items in the repertoires have scores
that are used to promote items which lead to successful imitations
and prune the other ones. In case of bad imitations, depending on the 
scores of the item used by the hearer, either this item is modified so as
to match better  the sound of the speaker, or a new item is created,
as close as possible to the sound of the speaker. 

This model is obviously very interesting since it was the first to show
a process of formation of vowel systems
within a population of agents (which was then extended to syllables
by Oudeyer in \citep{Oudeyer01b}). 
Yet, one has also to remark that the imitation game
that agents play is quite complex and requires a lot of 
assumptions about the capabilities of agents. 
From the description of the game, it is clear that to perform this kind
of imitation game, a lot of computational/cognitive power is needed.
First of all, agents need to be able to play a game, involving
successive turn-taking and asymmetric changing roles. Second, 
they need to have the ability to try to copy the sound production
of others, and be able to evaluate this copy. Finally, when they are
speakers, they need to recognize that they are being imitated
intentionally, and give feedback/reinforcement to the hearer 
about the success or not. The hearer has to be able to understand 
the feedback, i.e. that from the point of view of the other, 
he did or did not manage to imitate successfully. 

It seems that the level of complexity needed to form speech sound
systems in this model is characteristic of a society of agents which
has already some complex ways of interacting socially, and has already
a system of communication (which allows them for example to know
who is the speaker and who is the hearer, and which signal means
``good'' and which signal means ``bad''). The imitation game
is itself a system of conventions (the rules of the game), 
and agents communicate
while playing it. It requires the transfer of information from one agent to 
another, and so requires that this information be carried by some
shared ``forms''. So it pre-supposes that there is already
a shared system of forms. The vowel systems that appear
do not really appear ``from scratch''. This does not mean at all
that there is a flaw in de Boer's model, but rather that it
deals with the cultural evolution of speech rather than with the origins
(or, in other terms it deals with the formation of languageS - ``les
langues'' in French - rather than with the formation of language - ``
le langage'' in French). Indeed, de Boer presented interesting
results about sound change, provoked by stochasticity and 
learning by successive generations of agents. But the model
does not address the bootstrapping question: how
the first shared repertoire of forms appeared, in a society
with no communication and language-like interaction patterns?
In particular, the question of why agents imitate each other
in the context of de Boer's model (this is programmed in) is open.

The ``naming game'' described in \citep{Kaplan01}, or the ``iterated
learning model'' described in \citep{Kirby01}, are based on similar
strong assumptions concerning the cognitive capabilities of the agents.
\citep{Kaplan01} pre-supposes the capacity to play a game with rules
even more complex than in the ``imitation game''. \citep{Kirby01} pre-supposes
complex parsing capabilities as well as non-trivial generalization mechanisms
which seem to be language specific, even if its simulation does not use
an explicit functional pressure for communication. And, most importantly,
both pre-suppose the existence of a speech convention: their agents
can transmit and recognize ``labels'' or lists of letters directly (what they
learn is what these labels mean for the others in the case of \citep{Kaplan01},
or how these streams of letters are syntactically organized by the others in the case
of \citep{Kirby01}).

This shows that existing work relies on agents whose innate cognitive
capacities are already very complex and ``quasi-linguistic'', and which
possess already a number of conventions. So far, we do not
know how these capacities and these conventions, especially the speech convention, 
might have appeared if we do not pre-suppose that speech already exists. This is why we need either
to provide the explanation of their origins, or we need to provide 
a mechanism of the origins of speech which does not necessitate
them and relies on much simpler capacities whose origins we can
understand without pre-supposing the existence of speech.

We are going to present in this paper the second option: we will
build an artificial system that will put forward the idea that indeed,
much simpler mechanisms can account for the formation of shared
acoustic codes, which may later on be recruited for speech communication. 
This mechanism relies heavily on self-organization, in the same manner
as in the explanation of the hexagonal shape of honey bee's cells,
where the self-organization due to the physics of packed wax cells
does most of the job. Before presenting this artificial system, we
will briefly describe our methodology.

\section{The method of the artificial}

 The ``method of the artificial'' consists in building a society of formal agents \citep{Steels01}. The scientific logic is 
abductive. These agents are computer programs implemented in robots which possess for example an 
artificial vocal tract, an artificial ear, and artificial neural networks that connect them. These components 
are inspired by what we know of their human counterpart, but we do not necessarily try to reproduce
faithfully what we know of the human brain structures. 
We then study the dynamics resulting from their 
interactions, and we try to determine in what conditions they reproduce phenomena analogous to those 
of human speech. This does not aim to show directly  what were the mechanisms which gave rise to 
human speech, but the aim is to show what types of mechanisms are plausible candidates. The building of 
this artificial system provides constraints to the space of possible theories, in particular by showing 
examples of mechanisms which are sufficient, and examples of mechanisms which are not necessary
(e.g. we will show that imitation or feedback are not necessary to explain the formation of 
shared discrete speech codes). 

Some criticisms are sometimes put forward about this approach of the origins of language
through the building of artificial systems.
The opposition is often based on the argument that computer models are based
on strong assumptions which are remote from reality or very difficult to
validate or refute. This comes from a misunderstanding of the methodology and of
the aim of the researchers who build these artificial systems.
It must be stated clearly that this kind of computer simulation does not intend
to provide directly an explanation about the origins of some aspects of the human language.
Rather, they are used to organize the thinking and the conceptualisation
of the problematic of the origins of language, by shaping the search space
of possible theories. They are used to evaluate the internal coherence of
existing theories, and to explore new theoretical ideas. Then of course,
these computational models need to be extended and selected so as to fit the observations,
and become actual scientific hypotheses of the origins of language. But because
the phenomena involved in the origins of language are complex, we must
first develop and conceptualise our intuitions about the possible dynamics, 
before trying to formulate actual hypotheses. Building abstract computer simulations
is so far the best tool for this purpose.

Another opposition is the argument that says that too many aspects are
modelled at the same time, at the price of modelling each of them over
simplistically. This criticism is related to the first one. It should be
answered again that this might still be useful because of complexity: 
some phenomena are understandable only through the interactions of many components.
Yet, most research projects studying human speech focus on very
particular isolated components like the study of the electro-mechanical 
properties of the cochlea, the architecture of the auditory cortex, the
acoustics of the vocal tract, the systemic properties of vowels systems, etc.
Of course, having detailed knowledge and understanding of each of these
components is fundamental. But focusing on each of them individually
might prevent us from understanding major phenomena of speech and
language (and might possibly prevent us from understanding some of
the aspects of each module). It is necessary to study their interactions
in a parallel track. Because, it is practically impossible to incorporate
all the knowledge that we have of each component in a simulation, and
because for most of them there exist no real agreement on how 
they work, we can only use simplistic models so far. Besides the
fact that simulations incorporating the interactions of many components
can provide insights on the phenomena of speech, it is quite possible
that using these simplistic models might also shed light on the
functioning of some of the components by opening new conceptual 
dimensions and new experiments in vivo. In return, the simplistic models 
will then be made more realistic,  which will then help the understanding 
of components, forming a virtuous circle.

\section{The artificial system}

The system is a generalization of the one we described in \citep{Oudeyer01a},
which was used to model the phenomenon called the ``perceptual magnet effect''. 
It is based on the building of an artificial system, composed of
agents\footnote{The term 'agent' is used in artificial intelligence as an
abbreviation of 'artificial software agent', and denotes a software entity
which is functionally equivalent to a robot (this is like a virtual robot in
the virtual environment of the computer)} endowed with working models of the vocal tract, of the
cochlea and of some parts of the brain. The complexity and degree
of reality of these models can be varied to investigate which
aspects of the results are due to which aspects of the model. 

As explained in \citep{Oudeyer01a}, this system contains neural maps
which are similar to those used in \citep{Guenther96} and \citep{Damper00}.
What is different is that on the one hand, motor 
and perceptual neural maps are coupled so that the learning of sounds affects
the production of sounds, and on the other hand, these two other works used
single agents that learnt an existing sound system, while here we use
several agents that co-create a sound system.

\subsection{Overview}

Each agent has one ear which takes measures of the vocalizations that it
perceives, which are then sent to its brain. It also has a 
vocal tract, whose shape is controllable and is used to produce
sounds. Typically, the vocal tract and the ear define three spaces:
the motor space (which will be for example 3-dimensional in the
vowel simulations with tongue body position, tongue height and lip
rounding); the acoustic space (which will be 4-dimensional in the vowel
simulation with the first four formants) and the perceptual space
(which corresponds to the information the ear sends to the brain, and
will be 2-dimensional in the vowel simulations with the first formant and
the second effective formant).

\begin{figure}[t]
 \centerline{\includegraphics[width=0.8\linewidth]{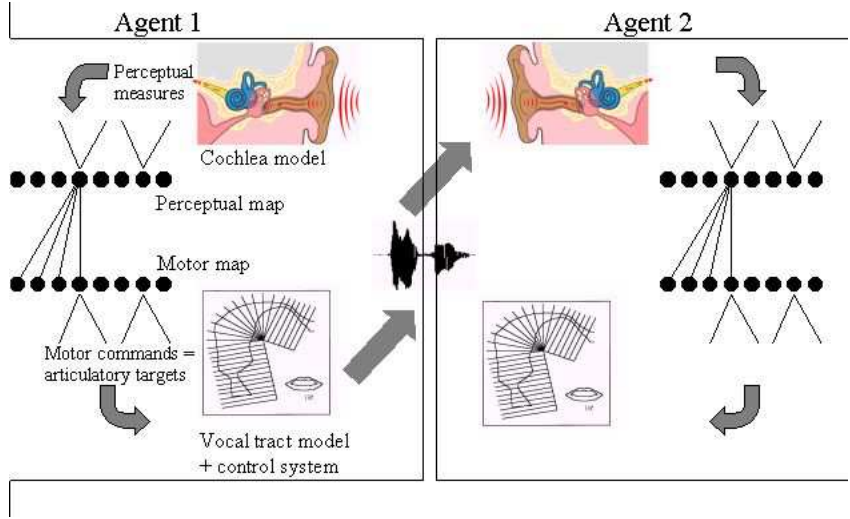}}
 \vskip 0.25cm
  \caption{\label{architecture}
           Architecture of the artificial system : agents are given an artificial ear, an artificial vocal tract, and an 
 artificial ``brain'' which couples these two organs. Agents are themselves coupled through their common 
 environment : they perceive the vocalizations of their neighbours.
 }
 \end{figure}

The ear and the vocal tract are connected to the brain, 
which is basically a set of interconnected artificial neurons
(the use of artificial neurons in computational models of the human
brain is described for example in \citep{Anderson95}). This
set of artificial neurons is organized into two neural topological maps: one perceptual
map and one motor map. 
Topological neural maps have been widely used for many models of cortical maps (\citep{Kohonen82}, 
\citep{Morasso98}),
which are the neural devices that humans   
have to represent parts of the outside world (acoustic, visual, touch etc.). 
Figure \ref{architecture} gives an overview of the architecture. 
We will now describe the technical details of the architecture.

\subsection{Motor neurons, vocal tract and production of vocalizations} 

\textbf{Structure.} A motor neuron $j$ is characterized by a preferred vector $v_{j}$ which
determines the vocal tract configuration which is to be reached
when it is activated and when the agent sends a GO signal to the
motor neural map. This GO signal is sent at random times by the
agent to the  motor neural map. As a consequence, the agent produces vocalizations
at random times, independently of any events.

\begin{figure}[t]
 \centerline{\includegraphics[width=0.8\linewidth]{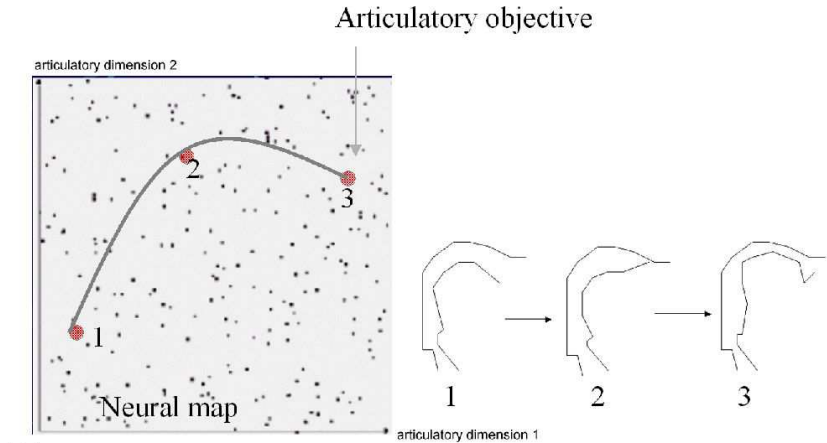}}
 \vskip 0.25cm
  \caption{\label{production}
          When an agent produces a vocalization, several motor neurons are activated in sequence.
Each of them corresponds to an articulatory configuration which has to be reached from
the current configuration. A sub-control system takes care of interpolating between the different
configurations.
 }
 \end{figure}

When an agent produces a vocalization, the neurons which are activated
are chosen randomly. Typically, 2, 3 or 4 neurons are chosen and activated
in sequence. Each activation of a neuron specifies, through its preferred
vector, a vocal tract configuration objective that a sub-system takes care of reaching
by moving continuously the articulators. In this paper, this sub-system is
simply a linear interpolator, which produces 10 intermediate configurations
between each articulatory objective, which is an approximation of a dynamic
continuous vocalization and that we denote $ar_{1}, ar_{2}, ..., ar_{N}$.
We did not use realistic mechanisms like the propagation
techniques of population codes proposed in \citep{Morasso98}, because these
would have been rather computationally inefficient for this kind of experiment. 
Figure \ref{production} illustrates this process in the case of the abstract
2-dimensional articulatory space that we will now describe. 

Indeed, the articulatory configurations will be coded in an abstract space in a first set
of simulations, and coded in a realistic space in a second more realistic set of simulations.
Also, in each case, we use an artificial vocal tract to compute an acoustic image
of the dynamic articulation.

In the abstract simulations, the articulatory configurations $ar_{i}=(d1_{i}, d2_{i})$ are just points in $[0,1]^{2}$.
The vocal tract is here a random linear mapping: in order to compute the acoustic
image of an articulatory trajectory defined by the sequence of articulations $ar_{1}, ar_{2}, ..., ar_{N}$,
we compute the trajectory of the acoustic images of each articulation in the acoustic space with
the formula: $$ac_{i} = (r_{1}.d1_{i} + r_{2}.d2_{i})/2$$ where $ac_{i}$ is the acoustic image
of $ar_{i}$ and  $r_{1}$ as well as $r_{2}$ are fixed random numbers. 

In the more realistic simulations, we use a vocal tract model of vowel production
designed by \citep{deBoer01}. We use vowel production only because there exists
this computationally efficient and rather accurate model, but one could do simulations
with a vocal tract model which models consonants if efficient ones were available.
The three major vowel articulatory parameters 
\citep{Ladefoged96} are used: lip rounding,  tongue height and tongue position.
The values within these dimensions are between 0
and 1, and a triplet of values $ar_{i}=(r, h, p)$ defines an articulatory configuration.
The acoustic image of one articulatory configuration is a point in the 4-dimensional
space defined by the first four formants, which are the frequencies of the peaks in the frequency spectrum,
and is computed with the formula :
\\ \\
$F_{1} = ((-392 + 392r)h^{2} + (596 - 668r)h + (-146 + 166r))p^{2}  
         + ((348 - 348r)h^{2} \\ + (-494 + 606r)h + (141 - 175r))p  
         + ((340 - 72r)h^{2} + (-796 + 108r)h \\ + (708 - 38r))$ \\ \\
$F_{2} =  ((-1200 + 1208r)h^{2} + (1320 - 1328r)h + (118 - 158r))p^{2}
         + ((1864 - 1488r)h^{2} \\ + (-2644 + 1510r)h + (-561 + 221r))p
         + ((-670 + 490r)h^{2} + (1355 - 697r)h \\ + (1517 - 117r)) $ \\ \\
$F_{3} =  ((604 - 604r)h^{2} + (1038 - 1178r)h + (246 + 566r))p^{2}
         + ((-1150 + 1262r)h^{2} \\ + (-1443 + 1313r)h + (-317 - 483r))p
         + ((1130 - 836r)h^{2} + (-315 + 44r)h \\ + (2427 - 127r)) $ \\ \\
$F_{4} =  ((-1120 + 16r)h^{2} + (1696 - 180r)h + (500 + 522r))p^{2}
         + ((-140 + 240r)h^{2} \\ + (-578 + 214r)h + (-692 - 419r))p
         + ((1480 - 602r)h^{2} + (-1220 + 289r)h \\ + (3678 - 178r)) $\\ \\
These were derived from polynomial interpolation based on a database
of real vowels presented in \citep{Vallee94}. Details are given in \citep{deBoer01}.

\textbf{Plasticity.} The preferred vector of each neuron in the motor map is updated
each time the motor neurons are activated (which happens both when
the agent produces a vocalization and when it hears a vocalization
produced by another agent, as we will explain below). This update
is made in two steps : 1) one computes which neuron $m$ is most activated
and takes the value $v_{m}$ of its preferred vector ; 2) the preferred vectors
of all neurons are modified with the formula:
\begin{center}
$v_{j,t+1} = v_{j,t} + 0.001.G_{j,t}(s).(v - v_{j,t})$
\end{center} 
where $G_{j,t}(s)$ is the activation of neuron $j$ at time t with the stimulus $s$ (as
we will detail later on) and $v_{j,t}$ denotes the value of $v_{j}$ at time $t$.
This law of adaptation of the preferred vectors has the consequence that the more a
particular neuron is activated, the more the agent will produce articulations
which are similar to the one coded by this neuron. This is because geometrically,
when $v_{m}$ is the preferred vector of the most active neuron,
the preferred vectors of the neurons which are also highly activated
are shifted a little bit towards $v_{m}$.
The initial value of all the preferred vectors of the motor neurons is random
and uniformly distributed. There are in this paper 500 neurons in 
the motor neural map (above a certain number of neurons, which is
about 150 in all the cases presented in the paper, nothing changes
if this number varies).

\subsection{Ear, perception of vocalizations and perceptual neurons}

We describe here the perceptual system of the agents, which is used when
they perceive a vocalization. As explained in the previous paragraphs, this
perceived vocalization takes the form of an acoustic trajectory, i.e. a
sequence of points which approximate the continuous sounds. In the abstract
simulations, these points are in the abstract 2-D space which we described
above. In this case, the acoustic space and the perceptual space are equal.
 In the more realistic simulations, these points are in the 4-D space 
whose dimensions are the first four formants of the acoustic signal.
In this case, we use also a model of our ear which transforms this 4-D
acoustic representation in a 2-D perceptual representation that we know is
close to the way humans represent vowels. This model was used in
 \citep{Boe95} and \citep{deBoer01}. It is based on the observations by
\citep{Carlson70} who showed that the human ear is not able to distinguish 
the frequency peaks with narrow bands in the high frequencies.
The first dimension is the first formant, and the second dimension is 
the second effective formant:
\begin{center}

\[ F_{2}^{'} = \left\{ \begin{array}{l}
                   F_{2},\  \mbox{\rm if} \  F_{3} - F_{2} > c \\
                   \frac{(2-w_{1})F_{2}+w_{1}F_{3}}{2},\ \mbox{\rm if} \ F_{3} -
                   F_{2} \leq c\ \mbox{\rm and}\ F_{4} - F_{2} \geq c \\
                   \frac{w_{2}F_{2} + (2-w_{2})F_{3}}{2} - 1,\ \mbox{\rm if} \  F_{4}
                   - F_{2} \leq c \ \mbox{\rm and}\  F_{3} - F_{2} \leq F_{4} - F_{3} \\
                   \frac{(2+w_{2})F_{3} - w_{2}F_{4}}{2} - 1,\ \mbox{\rm if} \ F_{4}
                   - F_{2} \leq c \ \mbox{\rm and}\  F_{3} - F_{2} \geq F_{4} - F_{3} 
                   \end{array}
                 \right. \]
\end{center}
with
$$w_{1} = \frac{c - (F_{3} - F{2})}{c}$$
$$w_{2} = \frac{(F_{4} - F_{3}) - (F_{3} - F_{2})}{F_{4} - F_{2}}$$
where $c$ is a constant of value 3.5 Barks.

In both cases (abstract and realistic simulations), the agent gets as input
to its perceptual neural system a trajectory of perceptual points. Each of 
these perceptual points is then presented in sequence to its perceptual
neural map (this models a discretization of the acoustic signal by the ear
due to its limited time resolution).

The neurons $i$ in the perceptual map have a gaussian tuning function
which allows us to compute the activation of the neurons upon the reception
of an input stimulus. 
If we denote by $G_{i,t}$ the tuning function of neuron $i$ at time $t$, 
$s$ is a stimulus vector, then the form of the function is:
\begin{center}
 $G_{i,t}(s) = \frac{1}{\sqrt{2\pi}\sigma}e^{-\frac{1}{2}
(v_{i,t}.s)^{2} / \sigma^{2}}$
\end{center}
where the notation $v_{1}.v_{2}$ denotes
the scalar product between vector $v_{1}$ and vector $v_{2}$,
and $v_{i,t}$ defines the center of the gaussian at time $t$ and is called
the preferred vector of the neuron. This means that when a 
perceptual stimulus is sent to a neuron $i$, then this neuron
will be activated maximally if the stimulus has the same value
as $v_{i,t}$.
The parameter $\sigma$ determines the width of the gaussian,
and so if it is large the neurons are broadly tuned (a value
of 0.05, which is used in all simulations here, means that a neuron responds substantially
to 10 percent of the input space). 

When a neuron in the perceptual map is activated because of a stimulus, 
then its preferred vector is changed.
The mathematical formula of the new tuning function is: 
 \begin{center}
 $G_{i,t+1}(s) = \frac{1}{\sqrt{2\pi}\sigma} 
                    e^{- \frac{1}{2} (v_{i,t+1}.s)^{2} / \sigma^{2}}$ 
\end{center}
where $s$ is the input, and $v_{i,t+1}$ the preferred vector
of neuron $i$ after the processing of $s$:
\begin{center}
 $ v_{i,t+1} = v_{i,t} + 0.001.G_{i,t}(s).(s - v_{i,t})$
\end{center}
This formula makes that the distribution of preferred vectors evolves
so as to approximate the distribution of sounds which are heard.

The initial value of the preferred vectors of all perceptual neurons
follows a random and uniform distribution. There are 500 neurons
in the perceptual map in the simulations presented in this paper.

\subsection{Connections between the perceptual map and the motor map}

Each neuron $i$ in the perceptual map is connected unidirectionally
to all the neurons $j$ in the motor map. The connection between
the perceptual neuron $i$ and the motor neuron $j$ is characterized
by a weight $w_{i,j}$, which is used to compute the activation of
neuron $j$ when a stimulus $s$ has been presented to the perceptual map, 
with the formula :
\begin{center}
$G_{j,t}(s) = \frac{1}{\sqrt{2\pi}\sigma} * 
                    e^{-\sum_{i} w_{i,j}G_{i,t}(s) / \sigma^{2}}$ 
\end{center}

The weights $w_{i,j}$ are initially set to a small random value, and evolve so as
to represent the correlation of activity between neurons. This is how agents will learn
the perceptual/articulatory mapping. The learning rule is hebbian \citep{Sejnowsky77}:
\begin{center}
$\delta w_{i,j} = c_{2}(G_{i} - < G_{i} >)(G_{j} - < G_{j} >)$
\end{center}
where $G_{i}$ denotes the activation of neuron $i$ and $< act_{i} >$ the
mean activation of neuron $i$ over a certain time interval
(correlation rule). $c_{2}$ denotes a small constant.
This learning rule applies only when the motor neural map is already activated before
the activations of the perceptual map have been propagated,
i.e. when an agent hears a vocalization produced by itself. This amounts to learning
the perceptual/motor mapping through vocal babbling. 

Note that this means that the motor neurons can be activated either
through the activation of the perceptual neurons when a vocalization is
perceived, or by direct activation when the agent produces
a vocalization (in this case, the activation of the chosen neuron is set
to 1, and the activation of the other neurons is set to 0).
Because the connections are unidirectional, the propagation
of activations only takes place from the perceptual to the articulatory map
(this does not mean that a propagation in the other direction would change
the dynamics of the system, but we did not study this variant).

This coupling between the motor map and the perceptual map has an important
dynamical consequence: the agents will tend to produce more
vocalizations composed of sounds that they have already heard. Said another
way, when a vocalization is perceived by an agent, this increases
the probability that the sounds that compose this vocalization will be re-used
by the agent in its future vocalizations. It is interesting to note that this phenomenon
of phonological attunement is observed in very young babies \citep{Vihman96}.

\subsection{Recurrence of the perceptual map and of the motor map}

Here we present an addition to the architecture presented in the previous 
paragraphs which is not crucial for the system\footnote{This means that
we do not need this addition for the main results of the system: 1) the formation
of shared discrete speech codes; 2) the formation of statistical regularities similar
to those of humans in the formed phonemic repertoires.} but which allows us both to
model the additional feature of categorization and to visualize the dynamics
of the rest of the system.  

This addition is based on the concept of population vector developed by \citep{Georgopoulos88},
and used in a similar setup by \citep{Guenther96}.
It proposes that the stimuli which are stored in the neural maps through the distributed activations
of neurons, can be decoded or re-constructed by some other parts of the brain by computing
the sum of all preferred vectors of the neurons weighted by their activity and normalized.
Technically, the population vector corresponding to the pattern of activations of the neurons $i$  
of a neural map when they have been activated by the stimulus $s$ is:
\begin{center}
$ pop(s) = \frac{\sum_{i} G_{i}(s)*v_{i}}{\sum_{i} G_{i}(s)} $ 
\end{center}
In general, $pop(s)$ is not exactly the same point as $s$, because this decoding scheme
is imprecise. But this imprecision can be exploited usefully.
Indeed, now we add a re-entrance of this re-constructed stimulus $pop(s)$: it is fed back
as an input to the neural map. And this gives rise to a new pattern of activations, which
is re-decoded, and the result is again fed back as input, and this is iterated until
a fixed point is reached. Indeed, this recurrent system has properties which can
be shown to be equivalent to Hopfield neural networks \citep{Hopfield82} and is very similar
to the system of \citep{Morasso98}: whatever the initial pattern of activations due to the 
perception of a stimulus, the cycle coding-decoding always converges on a fixed point 
(fixed pattern of activation). This process models a categorizing behaviour,
and the fixed point is the category which has been recognized by the system. Note that
this process is applied to each neural map only after all activations have been
propagated and after the learning rules have been applied (so this extension does
not modify the dynamics induced by the mechanisms presented in the previous
paragraphs).

A nice property is that the fixed pattern of activation when the system has converged
represents a particular stimulus which is basically the prototype of a category.
Indeed, if the preferred vectors of a neural map self-organize in clusters as will
happen in the simulations, each cluster, coding for a discrete ``mode'' or phoneme,
will have its center which coincides with the fixed point which is reached when
a stimulus close to this cluster is perceived. This center and associated fixed
point also represent the point of maximal density of neurons in the vicinity of
the cluster. More generally, the properties of this recurrent system make that its
dynamics is easy to represent. Indeed, if we use a stimulus space with 2 dimensions,
then each cycle coding/decoding can also be represented in 2-D : the input point
is represented by the beginning of an arrow, and the re-constructed point is
represented by the end of an arrow. By plotting all the arrows corresponding
to the first iteration of this process for all the points on a regular grid, we can
have a general view of the different basins of attractions and their fixed points,
which are at the same time the zones of maximal density of the clusters.
We will use this kind of plot to visualize the results of all our simulations, as
explained below.

\subsection{Coupling of agents}

The agents are put in a world where they move randomly. At random
times, a randomly chosen agent sends a GO signal and produces a vocalization. The agents
which are close to it can perceive this vocalization. Here, we fix
the number of agents who can hear the vocalization of another to 1 (we pick
the closest one). This is a non-crucial parameter of the simulations, since basically nothing
changes when we tune this parameter, except the speed of convergence
of the system (and this speed is lowest when the parameter is 1).
Technically, this amounts to having a list of agents, and in sequence 
picking up randomly two of them, have one produce a vocalization, and
the other hear it. Typically, there are 20 agents in the system. This is 
also a non-crucial parameter of the simulation : nothing changes except
the speed of convergence.

\subsection{What the system does not assume}

It is crucial to note that as opposed to most of simulations on the
origins of language that exist in the literature  \citep{Cangelosi02}),
our agents do not play here a ``language game'', in the sense that
there is no need to suppose an extra-linguistic protocol 
of interaction such as who should give feedback to whom and at what
particular moment and for what particular purpose. 
In particular agents do not play the ``imitation game'' which is for example used
in \citep{deBoer01}. Indeed, it is crucial to note that agents DO NOT
imitate each other in the simulations we present. Indeed, imitation
involves at least the reproduction of another's vocalization now or later:
here, one agent which hears another one never produces a vocalization
in response, and does not store the heard vocalization so as to reproduce
it later. The only consequence of hearing a vocalization is that it increases
the probability, for the agent which hears it, of producing later on vocalizations
whose parts are similar to those of the heard vocalization. Of course
it might happen, specially when the system has converged on a few
modes, that an agent produces a vocalization that it has already heard,
but this is no more an imitation than a human producing the same vowels
as another when responding to a question for example. 
The interactions of agents are not structured, 
there are no roles and no coordination. In fact, they have no social skill at all.
They do not distinguish between their own vocalizations
and those of others. They do not communicate. Here,
``communication'' refers to the emission of a signal by an individual
with the aim of modifying 
the state of at least one other agent,
which does not happen here. Indeed, agents do not even have means
to detect or represent other agents around them, 
so it would be difficult to say that they communicate. Finally, not only there are no
social force which act as a pressure to distinguish sounds, but there
are no internal force which act as a pressure to have a repertoire
of different discrete sounds : indeed, there are no repulsive forces
in the dynamics which update the preferred vectors of the neural maps.

\section{Dynamics}

We will study the dynamics of the artificial system in two different cases:
the first one is when the abstract linear articulatory synthesizer is used, while
the second one is when the realistic articulatory synthesizer is used.

\subsection{Using the abstract linear articulatory/perceptual mapping}

The present experiment used a population 
of 20 agents. Let us describe first what we obtain when agents use the abstract linear
articulatory synthesizer.

\begin{figure}[t]
\centerline{\includegraphics[width=0.75\linewidth]{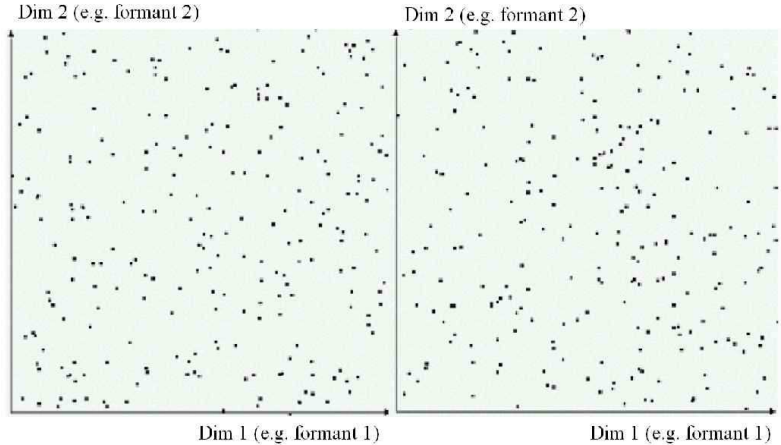}}
\vskip 0.25cm
 \caption{\label{initNeurons} Perceptual neural maps of two agents at the beginning
 (the two agents are chosen randomly among a set of 20 agents).
 Units are arbitrary. Each of both square represents the perceptual map of one agent.
}
\end{figure}

\begin{figure}[t]
\centerline{\includegraphics[width=0.75\linewidth]{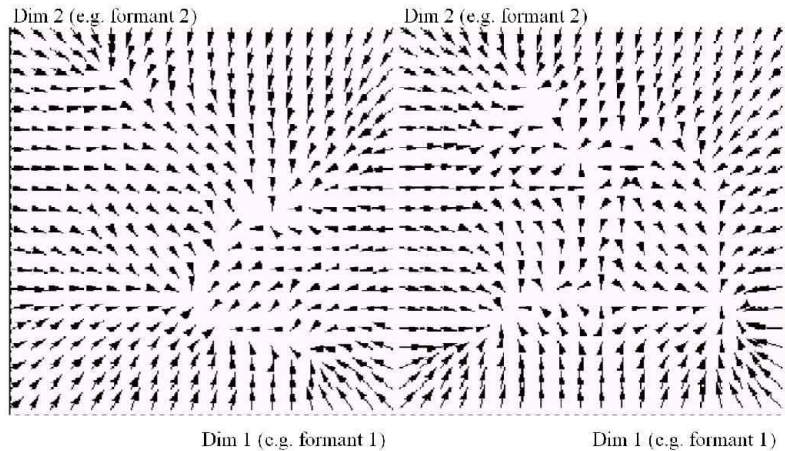}}
\vskip 0.25cm
 \caption{\label{initFields}
Representation of the same two agent's attractor field initially.
}
\end{figure}

Initially, as the preferred vectors of neurons are
randomly and uniformly distributed across the space, the different
targets that compose the vocalizations of agents are also randomly
and uniformly distributed. Figure \ref{initNeurons} shows the preferred vectors
of the neurons of the perceptual map of two agents. We see that
they cover the whole space uniformly. They are not organized.
Figure \ref{initFields} shows the basins of attraction associated with
the coding/decoding recurrent process that we described earlier.
The beginning of an arrow represents a pattern of activations at time
$t$ generated by presenting a stimulus whose coordinates
correspond to the coordinates of this point. 
The end of the arrow represents the pattern of activations of the neural map after one
iteration of the process. The set of all arrows provides a visualization of
several iterations: start somewhere on the figure, and follow the
arrows. At some point, for every initial point, you get to a fixed
point. This corresponds to one attractor of the network dynamic, 
and the fixed point to the category of the stimulus that gave
rise to the initial activation. The zones defining stimuli which
fall in the same category are visible on the figure, and are called
basins of attractions. With initial preferred vectors uniformly spread
across the space, the number of attractors as well as the boundaries
of their basins of attractions are random.

\begin{figure}[t]
\centerline{\includegraphics[width=0.75\linewidth]{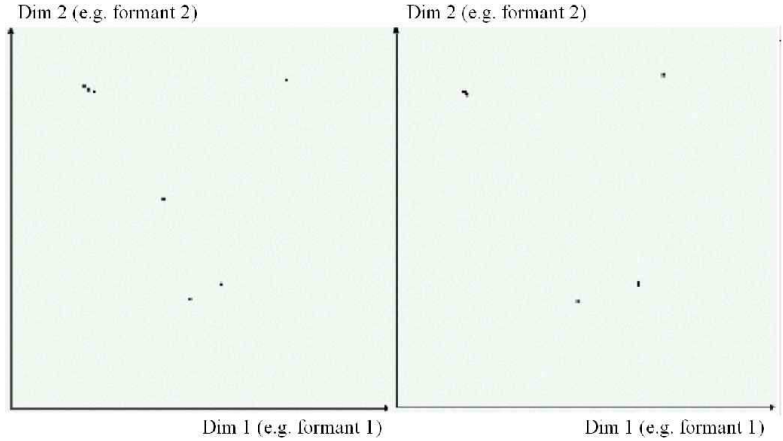}}
\vskip 0.25cm
 \caption{\label{finalNeurons}
   Neural maps after 2000 interactions,
   corresponding to the initial state of figure \ref{initNeurons}
   The number of points that one can see is fewer than the number
   of neurons, since clusters of neurons have the same preferred
   vectors and this is represented by only one point. 
}
\end{figure}

\begin{figure}[t]
\centerline{\includegraphics[width=0.75\linewidth]{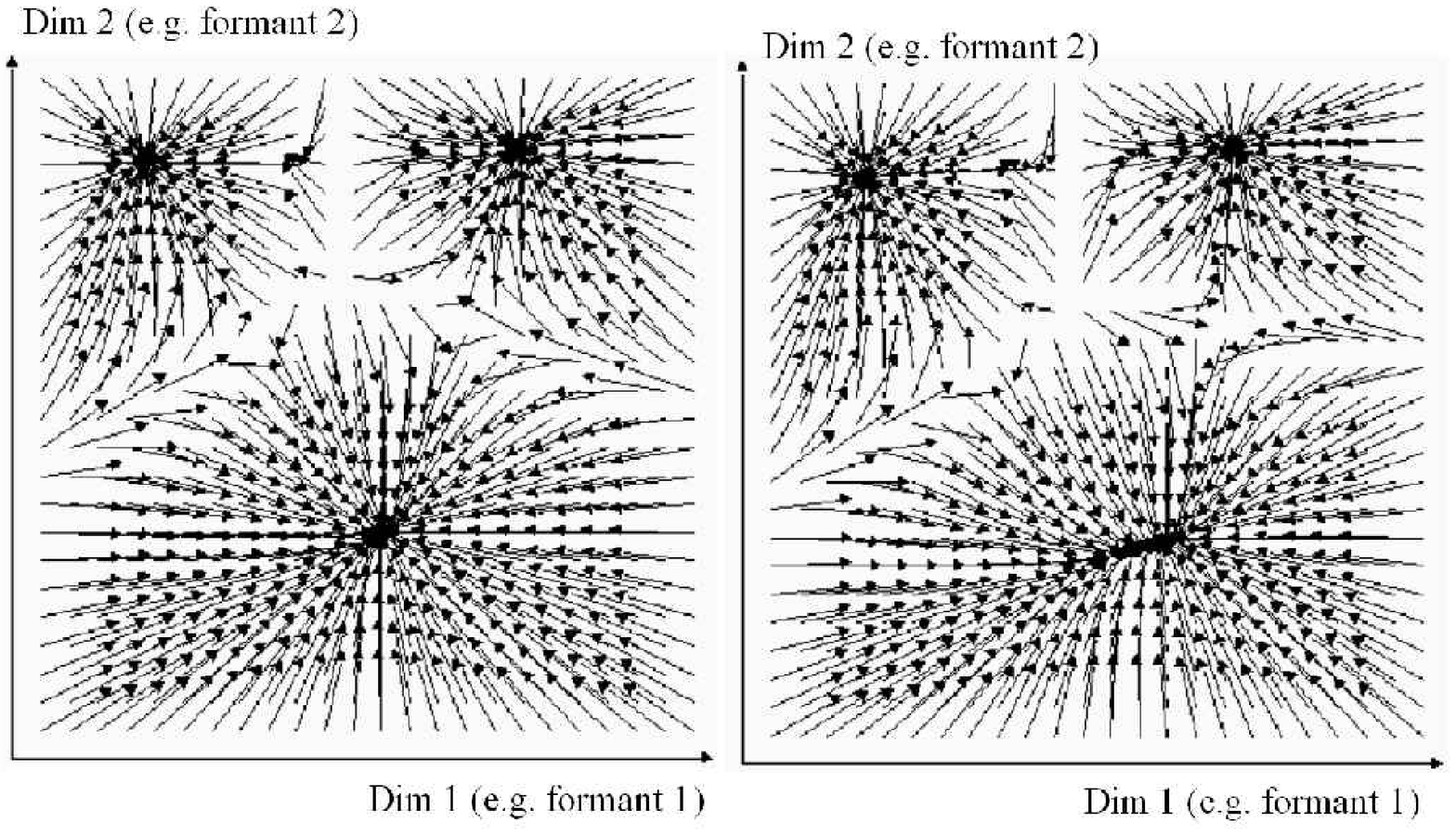}}
\vskip 0.25cm
 \caption{\label{finalFields}
   Representation of the attractor fields of 2 agents after
   2000 interactions. 
   The number of attractors is fewer that the number of points in the
   last figure. This is because in the previous figure, some points 
   corresponded to clusters and other to single points. The broad
   width of the tuning function makes that the landscape is smoothed
   and individual point which are not too far from clusters do not
   manage to form their own basin of attraction.
}
\end{figure}

The learning rule of the acoustic map is such that it evolves so as 
to approximate the distribution 
of sounds in the environment (but remember this is not due to imitation). 
All agents produce initially complex sounds composed of uniformly
distributed targets. Hence, this situation is in equilibrium.
Yet, this equilibrium is unstable, and fluctuations ensure that
at some point, the symmetry of the distributions of the produced sounds breaks: from time
to time, some sounds get produced a little more often than
others, and these random fluctuations may be amplified through
positive feedback loops. This leads to a multi-peaked distribution:
agents get in a situation like that of Figure \ref{finalNeurons}
which corresponds to Figure \ref{initNeurons} after
2000 interactions in a population of 20 agents. Figure \ref{finalNeurons}
shows that the distribution of preferred vectors
is no longer uniform but clustered (the same phenomenon happens in the motor
maps of the agents, so we represent here only the perceptual maps, as in the
rest of the paper). Yet, it is not so easy
to visualize the clusters with the representation in Figure \ref{finalNeurons},
since there are a few neurons which have preferred vectors not belonging
to these clusters. They are not statistically significant, but introduce
noise into the representation. Furthermore, in the clusters, basically
all points have the same value so that they appear as one point. Figure \ref{finalFields}
shows better the clusters using the attractor
landscape that is associated with them. We see that there are now
three well-defined attractors or categories, and that there
are the same in the two agents represented (they are also the
same in the 18 other agents in the simulation).
This means that the targets the agents use now belong to one
of several well-defined clusters, and moreover can be classified automatically
as such by the recurrent coding/decoding process of the neural map. 
The continuum of possible targets has been broken, sound
production is now discrete. Moreover, the number of clusters that
appear is low, which automatically brings it about that targets are
systematically re-used to build the complex sounds that agents
produce: their vocalizations are now compositional. 
All the agents share the same speech code in any one simulation. Yet, in each
simulation, the exact set of modes at the end is different.
The number of modes also varies with exactly the same
set of parameters. This is due to the inherent stochasticity
of the process. We will illustrate this later in the paper.

It is very important to note that this result of crystallization holds 
for any number of agents (experimentally), and in particular with only one agent
which adapts to its own vocalizations. This means that
the interaction with other agents (i.e. the social component)
is not necessary for discreteness and compositionality to arise. But what is
interesting is that when agents do interact, then they
crystallize in the same state, with the same categories. 
To summarize, there are so far two results in fact:
on the one hand discreteness and compositionality arise thanks to the coupling 
between perception and production within agents, on the other
hand shared systems of phonemic categories arise thanks to the 
coupling between perception and production across agents. 
  
We also observe that the attractors that appear
are relatively well spread across the space. The prototypes
that their centres define are thus perceptually quite distinct. In terms
of Lindblom's framework, the energy of these systems is high.
Yet, there was no functional pressure to avoid close prototypes.
They are distributed in that way thanks to the intrinsic
dynamics of the recurrent networks and their rather large tuning
functions: indeed, if two neuron clusters
just get too close, then the summation of tuning functions 
in the iterative process of coding/decoding smoothes their
distribution locally and only one attractor appears. 

\begin{figure}[t]
\centerline{\includegraphics[height=2.5in,width=3in]{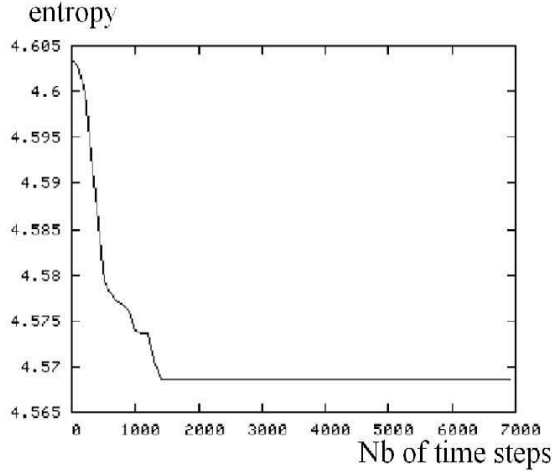}}
\vskip 0.25cm
 \caption{\label{entropy} Evolution of the entropy of the distributions of the preferred vectors
of the acoustic neurons of all agents.
}
\end{figure}

A last point to make is that what we call ``crystallization'' here is not
exactly a mathematical convergence, but a practical convergence of the
system. Indeed, as we explained in the previous sections, there are
only attractive forces that act on the preferred vectors of neurons.
No repulsive force is present. As a consequence, as these forces are
always strictly positive because of the gaussian tuning function, the point
of mathematical convergence of the system is when all preferred vectors
are clustered in one single point. Yet, this mathematical convergence
can not be reached in practice. Indeed, because we use a gaussian
tuning function, this attractive force becomes exponentially low as
stimuli get further from a given preferred vector. This has the consequence that
there is a first phase in the system during which a number of clusters
form, and sometimes ``melt'', until a state is reached in which the attraction
between clusters is so small that no new melting of clusters happens
before billions of time steps : in practice it is impossible to wait this
amount of time, which is much longer than the lifetime of agents.
This evolution can be illustrated by plotting the evolution of the
entropy of the distribution of the preferred vectors of all agents, 
as on Figure \ref{entropy}. We see a first phase of sharp decrease in the
entropy, and then a plateau. We use the term crystallization and
stop the simulations when this entropy plateau has been reached
(i.e. when the entropy value does not change for several thousands time steps).

Finally, it has to be noted that a crucial parameter of the simulation is the
parameter $\sigma$ which defines the width of the tuning functions. All
the results presented are with a value $0.05$. In \citep{Oudeyer03}, we present a study
of what happens when we tune this parameter. This study shows that the
simulation is quite robust to this parameter: indeed, there is a large zone
of values in which we get a practical convergence of the system in a state
where agents have a multi-peaked preferred vector distribution, as in the
examples we presented. What changes is the mean number of these peaks
in the distributions: for example, with $\sigma = 0.05$, we obtain between
3 and 10 clusters, and with $\sigma = 0.01$, we obtain between 6 and 15 clusters.
If $\sigma$ becomes too small, then the initial equilibrium of the system becomes
stable and nothing changes: agents keep producing inarticulate and holistic
vocalizations. If $\sigma$ is too large, then the practical convergence of the
system is the same as the mathematical convergence: only one cluster appears.

\subsection{Using the realistic articulatory/acoustic mapping}

In the previous paragraph, we supposed that the mapping from
articulations to perceptions was linear. 
In other words, constraints from the vocal
apparatus due to non-linearities were not taken into account. 
This was interesting
because it showed that no initial asymmetry in the system
was necessary to get discreteness (which is very asymmetrical).
In other words, this shows that there is no need to have
sharp natural discontinuities in the mapping from the
articulations to the acoustic signals and to the perceptions
in order to explain the existence of discreteness in speech sounds
(we are not saying that the non-linearities of the mapping do not
help, just that they are not necessary). 

Yet, this mapping has a particular
shape which introduces a bias into the pattern of speech sounds. 
Indeed, with the human vocal tract,
there are articulatory configurations for which a small change gives a small
change in the produced sound, but there are also articulatory configurations
for which a small change gives a large change in the produced sound.
While the neurons in the neural maps have initially
random preferred vectors with a uniform distribution, 
this distribution will soon become biased: 
the consequence of non-linearities will be that the learning
rule will have different consequences in different parts of the
space. For some
stimuli for which there are many articulatory configurations which produce
similar sounds, a lot of motor neurons will have their preferred vectors
shifted a lot, and for other stimuli, very few neurons will have
their preferred vectors shifted. This will very quickly lead
to non-uniformities in the distribution of preferred vectors in the
motor map, with more neurons
in the parts of the space for which small changes give
small differences in the produced sounds, and with fewer neurons
in the parts of the space for which small changes give
large differences in the produced sounds. As a consequence, the 
distribution of the targets that compose vocalizations will
be biased, and the learning of the neurons in the perceptual maps
will ensure that the distributions of the preferred vectors
of these neurons will also be biased.

\begin{figure}[h]
\centerline{\includegraphics[width=0.75\linewidth]{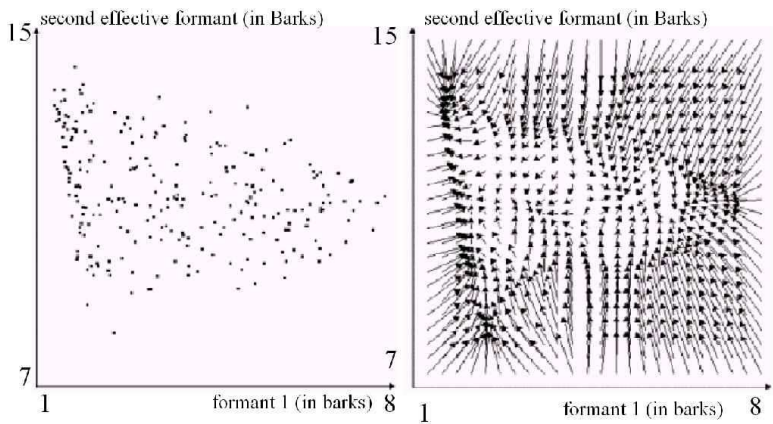}}
\vskip 0.25cm
 \caption{\label{initVowels}
   Neural map and attractor field of one agent within
   a population of 20 agents, after 200 interactions.
   Here the realistic articulatory synthesizer is used.
   The triangle which appears correspond to the so-called ``vocalic triangle'' \citep{Ladefoged96}.
 }
\end{figure}

We are going to study the consequence of using such a realistic
vocal tract and cochlear model in the system. We use the models
described earlier. To get an idea of the bias imposed by this mapping,
Figure \ref{initVowels} shows the state of the
acoustic neural maps of one agent after a few interactions (200) between
the agents.

\begin{figure}[!]
\centerline{\includegraphics[width=0.75\linewidth]{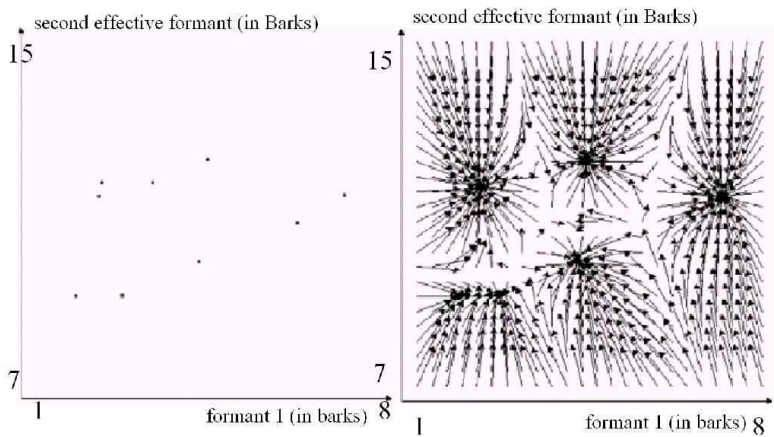}}
\vskip 0.25cm
 \caption{\label{finalVowels}
   Neural map and attractor field of the agent of figure \ref{initVowels}
   after 2000 interactions with other 20 agents. The corresponding
   figures of other agents are nearly identical. 
   The produced vowel system is here an instantiation the most frequent
   vowel system in human languages: /a, e, i, o, u/. 
}
\end{figure}

\begin{figure}[!]
\centerline{\includegraphics[width=0.75\linewidth]{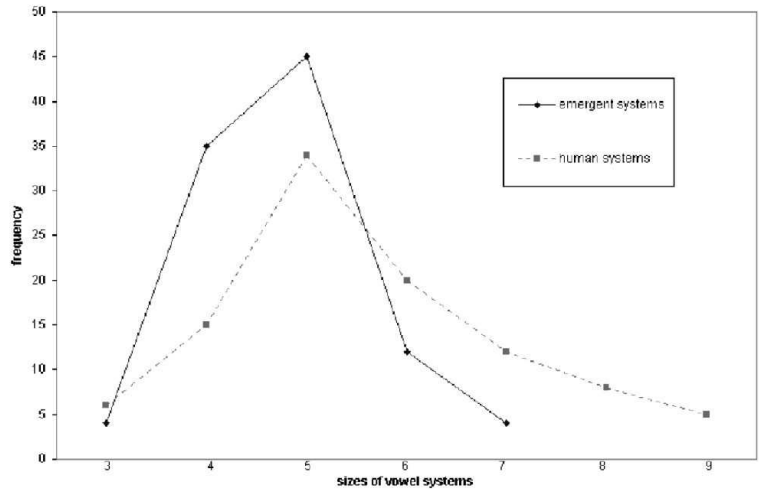}}
\vskip 0.25cm
 \caption{\label{distributions}
   Distribution of vowel inventories sizes in emergent and
   UPSID human vowel systems}
\end{figure}

\begin{figure}[!]
\centerline{\includegraphics[width=1\linewidth]{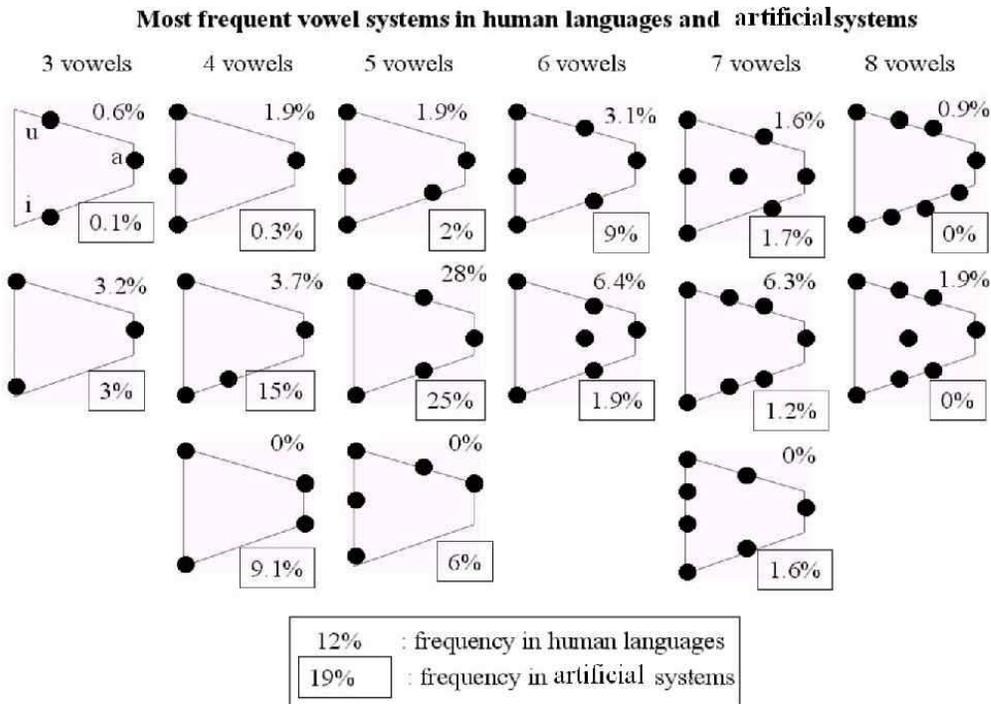}}
\vskip 0.25cm
 \caption{\label{inventories}
Distribution of vowel inventories structures in artificial and
UPSID human vowel systems. This diagram uses the same notations
than the one in \citep{Schwartz97}. Note that here, the
vertical axis is also F2, but oriented downwards.}
\end{figure}

A series of 500 simulations was run with the same set of 
parameters, and each time the number of vowels as well as the
structure of the system was checked. Each vowel system
was classified according to the relative position of the vowels,
as opposed to looking at the precise location of each of them.
This is inspired by the work of Crothers \citep{Crothers78} on universals
in vowel systems, and is identical to the type of classification
performed in \citep{deBoer01}. 
The first result shows that the distribution of vowel inventory sizes is
very similar to that of human vowel systems \citep{Ladefoged96}:
Figure \ref{distributions} shows the 2 distributions (in plain line the distribution
corresponding to the emergent systems of the experiment,
in dotted line the distribution in human languages), and in particular the fact
that there is a peak at 5 vowels, which is remarkable since
5 is neither the maximum nor the minimum number of vowels
found in human languages. The prediction made by the model
is even more accurate than the one provided by de Boer \citep{deBoer01} since
his model predicted a peak at 4 vowels. 
Then the structure of the emergent vowel systems was compared
to the structure of vowel systems in human languages as reported in \citep{Schwartz97}. 
More precisely, the distributions of structures in the 500 emergent systems
were compared to the distribution of structures in the 451 languages
of the UPSID database \citep{Maddieson84}. The results are shown in Figure \ref{inventories}. 
We see that the predictions are rather accurate, especially
in the prediction of the most frequent system for each size
of vowel system (less than 8). Figure \ref{finalVowels} shows an instance of the
most frequent system in both emergent and human vowel systems. 
In spite of the predictions of
one 4-vowel system and one 5-vowel system which appear 
frequently (9.1 and 6 percent of systems) in the simulations and never 
appear in UPSID languages, these results compare favourably to
those obtained in \citep{deBoer01}. In particular, we obtain
all this diversity of systems with the appropriate distributions
with the same parameters, whereas de Boer had to modify the level
of noise to increase the sizes of vowel systems. 
Yet, like de Boer, we are not able to predict systems with many
vowels (which are admittedly rare in human languages, but do exist).
This is certainly a limit of our model. Functional
pressure to develop efficient communication systems might be
necessary here.

\section{Conclusion}

This paper has presented a mechanism which provides a possible explanation
of how a speech code may form in a society of agents which do
not already possess means to communicate and coordinate in
a language-like manner (as opposed to the agents described in \citep{deBoer01}, \citep{Kaplan01} or
\citep{Oudeyer01b}), and which do not already possess a convention and complex
cognitive skills for linguistic processing (as opposed to the agents in \citep{Kirby01} for example).  
The agents in this paper have in fact no social skills at all. We believe that the value of the 
mechanism we presented resides in its quality of example of the kind of mechanism 
that might solve the language bootstrapping problem. We show how 
one crucial pre-requisite, i.e. the existence of an organized medium which 
can carry information in a conventional code shared by a population, may 
appear without linguistic features being already there.

The self-organized mechanism of this system appears as a necessary complement to the classical 
neo-Darwinian account of the origins of speech sounds. It is compatible with the classical neo-Darwinian 
scenario in which the environment favours the replication of individuals capable of speech. In this scenario,
our artificial system plays the same role as the laws of the physics of droplets in the explanation of the 
hexagonal shape of wax cells: it shows how self-organized mechanisms can facilitate the 
work of natural selection by constraining the shape space. Indeed, we show that natural selection did not necessarily
have to find genomes which pre-programmed the brain in precise and specific ways so as to 
be able to create and learn discrete speech systems. The capacity of coordinated social interactions
and the behaviour of imitation are also examples of mechanisms which are not necessarily pre-required
for the creation of the first discrete speech systems, as our system demonstrates. This 
draws the contours of a convincing classical neo-Darwinian scenario, by filling the conceptual gaps that
made it stay an idea rather than a real working mechanism.  

Furthermore, this same mechanism accounts for properties
of the speech code like discreteness, compositionality, universal
tendencies, sharing and diversity. We believe that this account
is original because: 1) only one mechanism is used to account for
all these properties and 2) we need neither a pressure for efficient
communication nor innate neural devices specific to speech (the
same neural devices used in the paper can be used to learn 
hand-eye coordination for example).
In particular, having made simulations both with and without non-linearities in the
articulatory/perceptual mapping allows us to say that in principle,
whereas the particular phonemes which appear in human languages 
are under the influence of the properties of this mapping, their mere existence,
which means the phenomenon of phonemic coding, does not require 
non-linearities in this mapping but can be due to the sensory-motor coupling dynamics.
This contrasts with the existing views that the existence of phonemic coding
necessarily need either non-linearities, as defended by \citep{Stevens72} or \citep{Carre88}, 
or an explicit functional pressure for efficient communication, 
as defended by \citep{Lindblom92}.

Models like the one of de Boer \citep{deBoer01} are to be seen
as describing phenomena occurring later in the evolutionary history
of language. More precisely, de Boer's model, as well as for
example the one presented in \citep{Oudeyer01b} for the formation of
syllable systems, deals with the 
recruitment of speech codes like those that appear
in this paper, and studies how they are further shaped and developed
under functional pressure for communication. Indeed,
if we have here shown that one can already go a long way without
such pressure, some properties of speech can only
be accounted for with it. An example is the existence of large
vowel inventories \citep{Schwartz97}.

\section{Acknowledgements}

I would like to thank Michael Studdert-Kennedy, Bart de Boer, Jim Hurford
and Louis Goldstein for their motivating feedback and their useful help
in the writing of this paper. 
I would also like to thank Luc Steels for supporting
the research presented in the paper. 

\bibliographystyle{elsart-harv}
\bibliography{jtpSelfOrganization}

\end{document}